\documentclass[letterpaper, 10 pt, journal, twoside]{IEEEtran}
\usepackage{amsmath,amsfonts}
\usepackage{bm}
\usepackage{algorithmic}
\usepackage{algorithm}
\usepackage{array}
\usepackage[caption=false,font=normalsize,labelfont=sf,textfont=sf]{subfig}
\usepackage{textcomp}
\usepackage{stfloats}
\usepackage{url}
\usepackage{verbatim}
\usepackage{graphicx}
\usepackage{cite}
\usepackage{hyperref}
\usepackage{stfloats}
\usepackage{tabularx}
\usepackage{booktabs}
\usepackage{multicol}
\usepackage{multirow}
\usepackage{bbding}
\usepackage{soul, color, xcolor}
\soulregister{\cite}{7}
\soulregister{\ref}{7}
\soulregister{\eqref}{7}
\soulregister{\textit}{7}
\usepackage{lineno}

\usepackage{hyperref}
\hypersetup{hidelinks}

\graphicspath{{figure/}}

\title{Data-Driven Dynamics Modeling of Miniature Robotic Blimps Using Neural ODEs With Parameter Auto-Tuning}

\begin{document}
% \pagewiselinenumbers
\switchlinenumbers	
\author{Yongjian Zhu, Hao Cheng, and Feitian Zhang
\thanks{Manuscript received: April, 29, 2024; Revised: July, 31, 2024; Accepted: October, 6, 2024. This paper was recommended for publication by Editor Aleksandra Faust upon evaluation of the Associate Editor and Reviewers’ comments. \it{(Corresponding author: Feitian Zhang.)}}
\thanks{The authors are with the State Key Laboratory of Turbulence and Complex Systems, and the Robotics and Control Laboratory, Department of Advanced Manufacturing and Robotics, College of Engineering, Peking University, Beijing, 100871, China (e-mail: {\tt\footnotesize yongjianzhu@pku.edu.cn}; {\tt\footnotesize h-cheng@stu.pku.edu.cn}; {\tt\footnotesize feitian@pku.edu.cn}).}
\thanks{Digital Object Identifier (DOI): see top of this page.}
}

\maketitle

\markboth{IEEE Robotics and Automation Letters. Preprint Version. Accepted October, 2024}
{Zhu \MakeLowercase{\textit{et al.}}: Data-Driven Dynamics Modeling of Miniature Robotic Blimps Using Neural ODEs With Parameter Auto-Tuning} 

\begin{abstract}
Miniature robotic blimps, as one type of lighter-than-air aerial vehicles, have attracted increasing attention in the science and engineering community for their enhanced safety, extended endurance, and quieter operation compared to quadrotors. Accurately modeling the dynamics of these robotic blimps poses a significant challenge due to the complex aerodynamics stemming from their large lifting bodies. Traditional first-principle models have difficulty obtaining accurate aerodynamic parameters and often overlook high-order nonlinearities, thus coming to their limit in modeling the motion dynamics of miniature robotic blimps. To tackle this challenge, this letter proposes the Auto-tuning Blimp-oriented Neural Ordinary Differential Equation method (ABNODE), a data-driven approach that integrates first-principle and neural network modeling. Spiraling motion experiments of robotic blimps are conducted, comparing the ABNODE with first-principle and other data-driven benchmark models, the results of which demonstrate the effectiveness of the proposed method.
\end{abstract}

\begin{IEEEkeywords}
Dynamics, calibration and identification, machine learning for robot control.
% robotic blimp, dynamics modeling, data-driven method, first-principle method
\end{IEEEkeywords}

\section{Introduction}
\IEEEPARstart{M}{iniature} robotic blimps are emerging as a promising technology with diverse applications including aerial photography \cite{10241589}, source localization \cite{10354821}, scientific education \cite{9062572} and human-robot interaction \cite{7989369}. These robotic blimps utilize lighter-than-air (LTA) gases for lifting, allowing for prolonged operation with minimal energy consumption. In addition, the flexibility of the blimp envelope contributes to enhanced safety for human-robot interaction \cite{8920528, lin2022miniature}.

Accurate dynamics modeling is critical to the motion control of robotic blimps. Traditionally, the first-principle method relies on domain knowledge and a deep understanding of the physical mechanisms governing robotic blimps to derive analytical models. For instance, Tao $\textit{et al.}$ developed a kinematic and dynamic model of a miniature autonomous blimp with a saucer-shaped envelope, simplified to three degrees of freedom (3-DOF) for ease of representation and identification \cite{tao2020modeling}. Azouz $\textit{et al.}$ employed unsteady potential flow analysis to estimate the added mass coefficients, leading to the formulation of a blimp dynamics model \cite{azouz2012computation}. Wang $\textit{et al.}$ characterized the movement of a blimp in the horizontal plane as a slider-like nonlinear system with uncertain disturbances \cite{wang2020disturbance}. In our previous research, we designed a robotic gliding blimp and established its 6-DOF dynamics model using Newton-Euler equations \cite{cheng2023rgblimp}. Due to their relatively large volume, robotic blimps exhibit complex, highly nonlinear, and non-negligible aerodynamic effects. A critical challenge lies in precisely identifying the parameters of the first-principle model. In addition, the first-principle model often falls short when robotic blimps execute agile and aggressive maneuvers, such as tight spirals with sharp turns, where traditional parameterized aerodynamics models fail to capture the high-order nonlinear dynamics involved \cite{bauersfeld2021neurobem}. 

Recent advancements in machine learning have expedited the efficient discovery of coherent patterns in complex data. Leveraging the capability of neural networks, data-driven methods have significantly outperformed traditional first-principle approaches in dynamics modeling for various aerial robotic systems with complex motion characteristics, e.g., quadrotors during agile and aggressive maneuvers \cite{loianno2016estimation,kaufmann2020deep, foehn2021time,savioloPhysicsInspiredTemporalLearning2022}. 

Aiming to precisely capture the complex dynamics of robotic blimps, this letter presents the ABNODE, a neural ordinary differential equation model coupled with a data-driven algorithm that automatically adjusts both the physical and network parameters of the dynamics model. Auto-tuning refers to the automatic adjustment of aerodynamic parameters in the first-principle module of the proposed ABNODE. This letter examines the spiral dynamics of robotic blimps, where the airflow field undergoes significant changes due to the blimp's movement, resulting in uncertainties in aerodynamic parameters and non-negligible high-order nonlinear residual dynamics.

The contribution of this letter is twofold. First, to the best of the authors' knowledge, ABNODE represents the first effort to establish the dynamics model of miniature robotic blimps by integrating first-principle and data-driven methods. ABNODE designs an automated training algorithm to refine both the physical and network parameters, accommodating uncertainties in the first-principle model and the high-order nonlinear residual dynamics. Second, this letter conducts extensive experimental validation and comparative studies of the proposed ABNODE against other baseline models. The experimental results demonstrate that ABNODE not only achieves more accurate dynamic motion prediction but also exhibits superior generalization capabilities. 

\section{Related Work}
\subsection{First-Principle Dynamics}
Taken as an example, the robotic gliding blimp \cite{cheng2023rgblimp} features a pair of fixed wings and a substantially larger envelope compared to typical quadrotors, potentially resulting in complex aerodynamics during its motion.

The blimp is modeled as a rigid-body system with six degrees of freedom. Table \ref{tab1} provides the reference frame and variable notations used in the dynamics model. Here, $\bm{x}=[\bm{p}^T,\bm{e}^T,\bm{v}_b^T,\bm{\omega}_b^T,\overline{\bm{r}}^T,\dot{\overline{\bm{r}}}^T]^T$ denotes the state vector, and $\bm{u}=[F_l^T,F_r^T,\overline{\boldsymbol{F}}^T]^T$ represents the control input vector, where $\overline{\bm{F}}=[F_x,F_y,F_z]^T=\overline{m}\ddot{\overline{\bm{r}}}$ denotes the force generated by the relative acceleration of the moving gondola with mass $\overline{m}$. 

\begin{table}[htbp]
  \caption{Nomenclature}
  \begin{center} 
  \begin{tabular}{m{2cm}m{6cm}}
    \hline \hline
    $O-xyz$                       & inertial frame \\
    $O_b-x_by_bz_b$             & body frame\\
    $O_v-x_vy_vz_v$             & velocity frame\\
    $m$                         & mass of blimp without gondola  \\
    $\overline{m}$              & mass of gondola                \\
    $\bm{p}        $            & position of blimp in $O$         \\
    $\bm{e}      $              & euler angles of blimp in $O$     \\
    $\bm{v}_{b}  $              & linear velocity of blimp in $O_b$  \\
    $\bm{\omega}_{b}$           & angular velocity of blimp in $O_b$ \\
    $\overline{\boldsymbol{r}}$ & centroid of gondola in $O$       \\ 
    $B$             & buoyancy of helium \\
    $F_l$           & left propeller thrust \\
    $F_r$           & right propeller thrust \\ 
    $\bm{R}$        & rotation matrix from $O$ to $O_b$\\
    $\bm{R}_{v}^{b}$ & rotation matrix from $O_b$ to $O_v$\\
    \hline \hline
  \end{tabular}
\end{center}
\label{tab1}
\end{table}
    % $d$           & distance between two propellers \\
 % $\bm{J}$        &  transformation of angular velocity from $O$ to $O_b$ \\ 
The first-principle dynamics model \cite{cheng2023rgblimp} is typically expressed as  
\begin{align}
    \left[\; \begin{array}{cc}
  \dot{\boldsymbol{p}} \\
  \dot{\boldsymbol{e}} \\
  \dot{\overline{\boldsymbol{r}}}
  \end{array} \; \right]=&
  \left[\begin{array}{cc}
  \boldsymbol{R} \boldsymbol{v}_b \\
  \boldsymbol{J} \boldsymbol{\omega}_b \\
  \dot{\overline{\boldsymbol{r}}}
  \end{array}\right] \label{eq:full1}\\ 
    \left[\begin{array}{cc}
     \dot{\boldsymbol{v}}_b \\
     \dot{\boldsymbol{\omega}}_b \\
     \ddot{\overline{\boldsymbol{r}}}
     \end{array}\right]=& \boldsymbol{A}\left[\begin{array}{cc}
     \tilde{\boldsymbol{f}} \\
     \tilde{\boldsymbol{t}} \\
     \mathbf{0}_{3 \times 1}
\end{array}\right]+\boldsymbol{B}\left[\begin{array}{cc}
     F_l \\
     F_r \\
     \overline{\boldsymbol{F}}
     \end{array}\right]
     \label{eq:full2}
\end{align}
Here, $\bm{A}$ and $\bm{B}$ are state-dependent matrices. $\tilde{\boldsymbol{f}}$ and $\tilde{\boldsymbol{t}}$ are calculated as

\begin{align}
       \tilde{\boldsymbol{f}}= & (m+\bar{m}) \boldsymbol{v}_b \times \boldsymbol{\omega}_b+\left(\boldsymbol{\omega}_b \times \boldsymbol{l}_g\right) \times \boldsymbol{\omega}_b+ \label{eq:ff} \\ \notag
  & (m g+\bar{m} g-B) \boldsymbol{R}^{\mathrm{T}} \boldsymbol{k}+\boldsymbol{F}_{\text {aero}}+2 \bar{m} \dot{\boldsymbol{r}} \times \boldsymbol{\omega}_b \\
   \tilde{\boldsymbol{t}}= & \boldsymbol{l}_g \times\left(\boldsymbol{v}_b \times \boldsymbol{\omega}_b\right)+\left(\boldsymbol{I}-\bar{m}\left(\overline{\boldsymbol{r}}^{\times}\right)^2\right) \boldsymbol{\omega}_b \times \boldsymbol{\omega}_b+ \label{eq:tt}\\ \notag
   &\boldsymbol{l}_g \times g \boldsymbol{R}^{\mathrm{T}} \boldsymbol{k}+\boldsymbol{T}_{\text{aero}}+2 \bar{m} \overline{\boldsymbol{r}} \times\left(\dot{\overline{\boldsymbol{r}}} \times \boldsymbol{\omega}_b\right) 
\end{align}
Here, $\bm{l}_g=m\bm{r}+\bar{m}\bar{\bm{r}}$ and $\bm{l}_g/(m+\bar{m})$ represents the centroid of the robot. $\bm{k}=[0,0,1]^T$ is the unit vector along the $Oz$ axis. \par
The aerodynamic forces $\bm{F}_\text{aero}$ and moments $\bm{T}_\text{aero}$ are expressed in the body frame and calculated as

\begin{align}
  \bm{F}_{\text{aero}}=& \bm{R}_v^b[-F_1,\;F_2, -F_3]^{T} \label{eq:Fa} \\
  \bm{T}_{\text{aero}}=& \bm{R}_v^b[\; M_1,\;M_2,\;M_3]^{T} \label{eq:Ta} 
\end{align}
where $F_1$, $F_2$, and $F_3$ represent the drag force, the side force, and the lift force, respectively. $M_1$, $M_2$, and $M_3$ represent the rolling moment, the pitching moment, and the yaw moment, respectively. These aerodynamic forces and moments are typically modeled as functions of the angle of attack $\alpha$, the sideslip angle $\beta$, and the velocity magnitude $V$, i.e.,
\begin{align}
   F_i =& 1 / 2 \rho V^2 A \;C_{F_i}(\alpha, \beta) \label{eq:Fi} \\
   M_i =& 1 / 2 \rho V^2 A \;C_{M_i}(\alpha, \beta)+K_i \bm{\omega}_b^{i}\;(i=1,2,3) \label{eq:Mi}
\end{align}
where $A$ represents the reference area, determined by the blimp design. $K_i$ denotes the rotation damping coefficients. The terms in the form of $C_{\square}(\alpha,\beta)$ denote the polynomials of $\alpha$ and $\beta$ with the orders of no more than four \cite{cheng2023rgblimp}.

Identifying the parameters in Eqs. (\ref{eq:Fi}) and (\ref{eq:Mi}) is crucial for accurately modeling robotic blimps \cite{tao2018parameter,ko2007gaussian}. Traditionally, these physical parameters are derived through polynomial regression of experimental trajectory data obtained from selected motions such as straight-line flight or spiraling flight, with either constant or time-varying speeds. It is noteworthy that measuring these physical aerodynamic parameters via wind tunnel experiments is challenging and costly due to the small size and the flexible design of the envelope. 
Additionally, first-principle modeling tends to overlook high-order nonlinear dynamics\cite{hu2020neural, klein1989estimation}, resulting in a parameterized model that lacks sufficient accuracy, especially during complex agile aerial maneuvers. 
\subsection{Data-Driven Modeling}
We rewrite the dynamics model (Eqs. \eqref{eq:full1}--\eqref{eq:Mi}) of the robotic blimp in a more compact form, i.e.,
\begin{align}
    \dot{\bm{x}}= & \bm{f}(\bm{x},\bm{u};\bm{\eta}) \label{eq: general} 
\end{align}
where $\bm{x}$ and $\bm{u}$ represent the system states and the control inputs, respectively. $\bm{\eta}=[\bm{\eta}_{F_i},\bm{\eta}_{M_i},\bm{\eta}_{K_i}]^{T}$ where  $i=1,2,3$ represents the overall uncertain physical parameters of the model. $\bm{\eta}_{F_i}$ and $\bm{\eta}_{M_i}$ represent the aerodynamic parameters in polynominals $C_{F_i}$ and $C_{M_i}$ in Eq. \eqref{eq:Fi} and Eq. \eqref{eq:Mi}, respectively. $\bm{\eta}_{K_i}$ represent the same damping parameters $K_i$ as in Eq. \eqref{eq:Mi}. The aim is to establish a continuous-time function, denoted as $\bm{f}$, to accurately model the dynamics given experimental flight data of the robotic blimp. \par
% There has been much research on establishing a continuous dynamics model for robots. 
In recent years, interdisciplinary research efforts spanning dynamics, robotics, machine learning, and statistics have led to the development of data-driven algorithms for modeling complex dynamical systems. For instance, Bruton $\textit{et al.}$ introduced the sparse identification of nonlinear dynamics with control (SINDYc), which utilized regression methods to select nonlinear terms from a predefined library and establish ODE models for dynamical systems \cite{kaiser2018sparse}. Bansal $\textit{et al.}$ employed a shallow feedforward neural network (FNN) to learn the dynamics model of a quadrotor \cite{7798978}. While FNNs are capable of modeling highly complex phenomena, they may struggle to capture time-correlated features. On the other hand, recurrent neural networks (RNN) are specialized architectures for building time-series models. However, RNNs encounter challenges such as the vanishing or exploding gradient problem and difficulty processing long sequences, rendering them challenging to train and unsuitable for robotic applications \cite{pmlr-v28-pascanu13}.\par
Taking the advantages of both domain knowledge and data-driven modeling techniques, researchers have proposed various physics-informed neural networks for modeling the dynamics of complex systems. For example, Saviolo $\textit{et al.}$ introduced the physics-inspired temporal convolutional network (PI-TCN), which initially aligned with experimental data distribution and subsequently with first-principle model predictions in a two-phase process \cite{savioloPhysicsInspiredTemporalLearning2022}. Jiahao $\textit{et al.}$ developed the knowledge-based neural ordinary differential equation (KNODE)\cite{jiahao_knowledge-based_2021}, which integrates neural ordinary differential equations \cite{chen2018neural} with first-principle modeling. \par
While PI-TCN and KNODE utilize both first-principle models and neural networks for dynamics modeling of robotic systems, they have limitations, especially concerning complex blimp dynamics. First, the model accuracy heavily depends on the initial setting of physical parameters. Inappropriate settings result in slow convergence or sometimes model failure. 
Second, these models typically learn system dynamics without differentiating between robot kinematics and dynamics. For blimps, the kinematics model is often accurate, while dynamics involve aerodynamic uncertainties overlooked by PI-TCN and KNODE. ABNODE is particularly effective for: 1) establishing continuous-time ODE models; 2) incorporating dynamics with uncertain physical parameters; and 3) utilizing accurate robot kinematics.

\section{Design of the ABNODE Model}
\label{section: Methodology}
\subsection{First-Principle and Neural Network Modules}
The ABNODE model consists of two modules including the first-principle module which incorporates the physical parameters of robotic blimps, and the neural network module, designed to capture residual dynamics. Given the accurate kinematics description (Eq. \eqref{eq:full1}) of the robotic blimp, the proposed ABNODE model only updates the dynamics (Eq. \eqref{eq:full2}), leaving the kinematics unchanged. Following the literature \cite{mohajerin2018deep,chee_knode-mpc_2022,duan2024machine} and assuming an additive form of high-order nonlinear uncertainties, we write the dynamics of the robotic blimp as
\begin{equation}
  \label{eq:AutoKNODE3}
  \dot{\bm{x}}=\bm{f}_{\rm{phy}}(\bm{x},\bm{u};\bm{\eta})+\Delta_{\rm{nn}}(\bm{x},\bm{u};\bm{\theta}) 
\end{equation}
where $\bm{f}_{\rm{phy}}$ represents the first-principle module derived in Eqs. (\ref{eq:full1}) and (\ref{eq:full2}). The vector $\bm{\eta}$ denotes the corresponding aerodynamics parameters. $\Delta_{\rm{nn}}$ represents the neural network module used to model uncertainty dynamics dependent on states and controls. The vector $\bm{\theta}$ represents network parameters. Identifying parameters $\bm{\eta}$ and $\bm{\theta}$ of the ABNODE model characterizes the dynamics modeling process.

\subsection{Physics-informed Loss}
Learning robot dynamics purely from data often leads to models with poor generalizability outside the training data distribution. The first-principle module offers insights into the dynamical system and contributes to accurate modeling if leveraged effectively. When the physical parameters of the first-principle module are inaccurate, network training tends to struggle to achieve convergence. Inspired by previous works \cite{savioloPhysicsInspiredTemporalLearning2022, baierHybridPhysicsDeep2022a,wang2021bridging}, this letter designs a physics-informed loss for training the ABNODE model, i.e., 
\begin{equation}
  \mathcal{L}_{\rm{ABNODE}}(\bm{\eta},\boldsymbol{\theta})=\bm{b}\mathcal{L}_{\rm{phy}}+(\neg \bm{b}) \mathcal{L}_{\rm{BNODE}} \label{eq:loss}
\end{equation} 
where $\mathcal{L}_{\rm{phy}}$ represents the first-principle loss used for obtaining proper physical parameters $\bm{\eta}$. $\mathcal{L}_{\rm{BNODE}}$ denotes the neural network loss for updating network parameters $\bm{\theta}$. Both the physical parameters $\bm{\eta}$ in the neural network module and the corresponding parameters in the first-principle module are updated simultaneously. The hyperparameter $\bm{b}$ is a boolean variable, where $\bm{b} \in \{0,1\}$. Symbol $\neg$ denotes logic negation, a fundamental operation in logic and mathematics, used for reversing the truth value of $\bm{b}$, such that $(\neg0)=1$ and $(\neg1)=0$. When the physical parameters are deemed unreliable, $\bm{b}$ is set to 1 to optimize the first-principle module, focusing on the physical parameters of the robotic blimp. Conversely, when confidence in the physical parameters is high, $\bm{b}$ is set to 0 to allow the network to fully explore and capture residual dynamics. 

Without loss of generality, this letter selects the widely-adopted mean square error as the loss function to quantify the optimization cost, i.e., 
\begin{equation}
    \mathcal{L}=\frac{1}{N-1}\sum_{i=2}^N\frac{1}{n}\Vert \bm{W}(\bm{x}_\text{p}(t_i)-\bm{x}(t_i))\Vert_2^2
    \label{eq:allloss}
\end{equation}
Here, $n$ represents the dimension of the system states. $\bm{x}(t_i)$ denotes the experimental value of the state vector at time $t_i$, while $\bm{x}_\text{p}(t_i)$ denotes the predicted value of the state calculated by the dynamics model. $\bm{W}\in\mathbb{R}^{n}\times\mathbb{R}^{n}$ is a diagonal matrix, defining the weights of different states. $N$ denotes the length of the experimental data. In the loss function $\mathcal{L}_{\rm{phy}}$, $\bm{x}_\text{p}(t_i)$ is represented as $\tilde{\bm{x}}(t_i)$, which denotes the output of the first-principle module. In loss $\mathcal{L}_{\rm{BNODE}}$, $\bm{x}_{\rm{p}}$ is represented as $\hat{\bm{x}}(t_i)$, which denotes the output of the neural network module. 
Specifically, $\tilde{\bm{x}}(t_i)$ and $\hat{\bm{x}}(t_i)$ are defined as 
\begin{align}
  & \tilde{\bm{x}}(t_{i+1})=\bm{x}(t_i)+\int_{t_i}^{t_{i+1}}\bm{f}_{\rm{phy}}(\bm{x},\bm{u};\bm{\eta})\mathrm{d}t \label{eq:x_tilde} \\
  & \hat{\bm{x}}(t_{i+1})=\bm{x}(t_i)+\int_{t_i}^{t_{i+1}}\bm{f}_{\rm{phy}}(\bm{x},\bm{u};\bm{\eta})+\Delta_{\rm{nn}}(\bm{x},\bm{u};\bm{\theta})\mathrm{d}t \label{eq:x_hat}
\end{align}
\subsection{Two-phase Model Training}
\begin{figure}[!t]
    \vspace{-2mm}
    \centerline{\includegraphics[width=9cm]{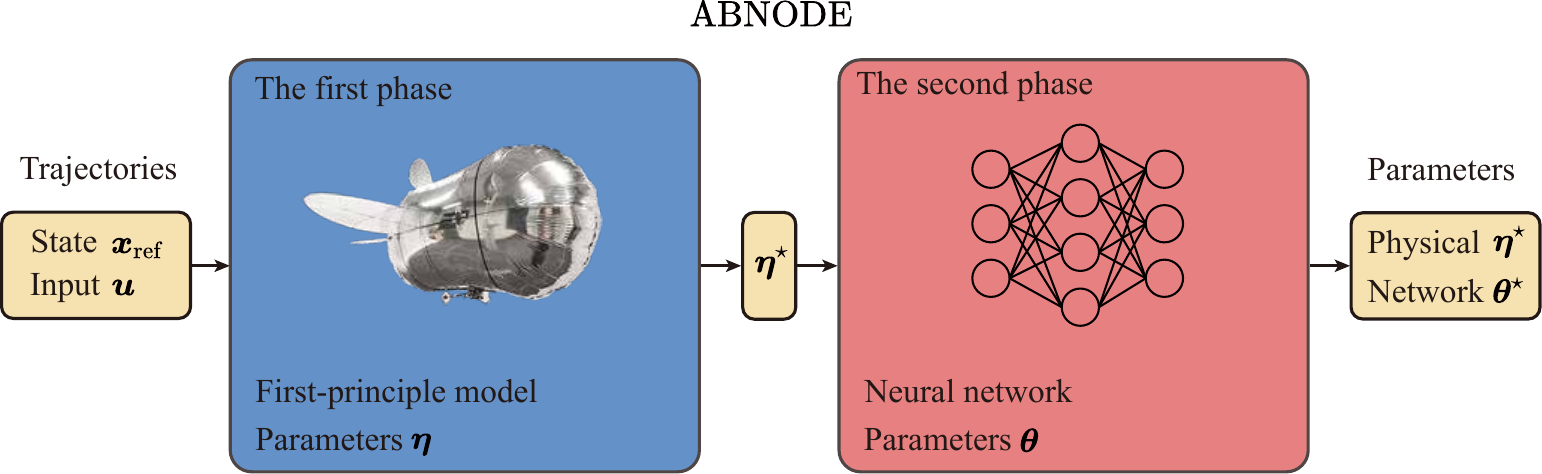}}
    \caption{An overview of the two-phase ABNODE model training strategy. The first phase optimizes the physical parameters $\bm{\eta}$ of the first-principle model while the second phase focuses on network parameters $\bm{\theta}$.}
    \label{fig:schematic}
    \vspace{-3mm}
\end{figure}

ABNODE features both trainable physical and network parameters along with a physics-informed loss function. This letter adopts a two-phase model training strategy, illustrated in Fig.~\ref{fig:schematic}. In the initial phase, $\bm{b}$ is set to 1 to prioritize updating the physical parameters $\bm{\eta}$, leveraging the potential of the first-principle module. Once convergence of the initial values of the physical parameters $\bm{\eta}^{\star}$ is achieved, $\bm{b}$ is then set to 0 to prioritize the updating of the neural network parameters $\bm{\theta}$. The proposed ABNODE is detailed in Algorithm~\ref{alg:AutoKNODE1}. $\bm{X}=[\bm{x}_0,\bm{x}_1,\cdots,\bm{x}_N]$ represents the experimental data of system states from time step $0$ to time step $N$. $\bm{\Upsilon}=[\bm{u}_0, \bm{u}_1,\cdots,\bm{u}_N]$ denotes the control inputs during the experiment. $\tilde{\bm{X}}=[\tilde{\bm{x}}_0,\tilde{\bm{x}}_1,\cdots,\tilde{\bm{x}}_N]$ and $\hat{\bm{X}}=[\hat{\bm{x}}_0,\hat{\bm{x}}_1,\cdots,\hat{\bm{x}}_N]$ represent the predicted states in the first phase and the second phase, respectively.

\subsubsection{First phase}
In this phase, the hyperparameter $\bm{b}=1$. The first-principle model $\bm{f}_{\rm{phy}}$ takes the experimental data of states $\bm{x}(t_i)$ and control $\bm{u}(t_i)$ as input and outputs the time-derivative of the states. Both $\bm{f}_{\rm{phy}}$ and the experimental data are passed to the ODE solver. Following the Runge-Kutta method, the predicted state at the next time step $\tilde{\bm{x}}(t_{i+1})$ is generated following Eq.~\eqref{eq:x_tilde}. The predicted state $\tilde{\bm{x}}(t_{i+1})$, combined with the control input $\bm{u}(t_{i+1})$, serve as the input to the first-principle module to calculate the time-derivative of the state at time $t_{i+1}$. This iterative process continues until the robot trajectory is determined. After $N-1$ time steps, we calculate the total loss $\mathcal{L}_{\rm{phy}}(\bm{\eta})$ and use the adjoint sensitivity method \cite{adjoint1962} to estimate the gradient for updating the physical parameters $\bm{\eta}$. After the first phase, the physical parameters $\bm{\eta}^{\star}$ are passed to the neural network module for the second-phase model training. \par

\subsubsection{Second phase}
In the second phase, with the hyperparameter $\bm{b}=0$, the focus shifts to training the neural network to model the residual dynamics. Both the first-principle and neural network modules contribute to calculating the time derivative of the state, with their outputs combined. At each time step, the predicted state $\hat{\bm{x}}(t_{i+1})$ and the control input $\bm{u}(t_{i+1})$ are passed to the first-principle and neural network modules to compute the time derivative of the states. This iterative process continues until the robot trajectory is determined. Subsequently, we compute the loss function $\mathcal{L}_{\rm{BNODE}}(\bm{\eta}^{\star},\bm{\theta})$ and update the network parameters accordingly. Thanks to the first phase, the training of the neural network module avoids significant divergence caused by inaccurate physical parameters. \par

\begin{figure}[!h]
\vspace{-5mm}
  \renewcommand{\algorithmicrequire}{\textbf{Input:}}
  \renewcommand{\algorithmicensure}{\textbf{Output:}}
  \begin{algorithm}[H]
    \caption{ABNODE}
    \label{alg:AutoKNODE1}
    \begin{algorithmic}[1]
        \REQUIRE State matrix $\bm{X}$, Control input matrix $\bm{\Upsilon}$, Dynamics model $\bm{f}_{\rm{phy}}$, First phase epochs $N_1$, Second phase epochs $N_2$  
        \ENSURE Physical parameters $\bm{\eta}$, Network parameters $\bm{\theta}$
        \STATE $k \gets 0$
        \STATE $\bm{\eta} \gets \bm{\eta}_0, \bm{\theta} \gets \bm{\theta}_0$
        \WHILE{$k<N_1+N_2$}
        \STATE Solve Eq. \eqref{eq:AutoKNODE3} with the ODE solver
        \IF{$k < N_1$}
        \STATE Generate the prediction $\tilde{\bm{X}}$ by Eq. \eqref{eq:x_tilde}
        \ELSE
        \STATE Generate the prediction $\hat{\bm{X}}$ by Eq. \eqref{eq:x_hat}
        \ENDIF
        \STATE Minimize the physical-informed loss \eqref{eq:loss}
        \IF{$k < N_1$}
        \STATE calculate the gradient of $\bm{\eta}$, $\bm{\eta} \gets\bm{\eta}_{k+1}$
        \ELSE
        \STATE calculate the gradient of $\bm{\theta}$, $\bm{\theta} \gets\bm{\theta}_{k+1}$
        \ENDIF
        \STATE $k \gets k+1$
        \ENDWHILE
    \end{algorithmic}
  \end{algorithm}
  \vspace{-5mm}
\end{figure}

\section{Experiment}
\label{section: Experiment}

\subsection{Experimental Setup}
\begin{figure}[htbp]
    \centerline{\includegraphics[width=8.8cm]{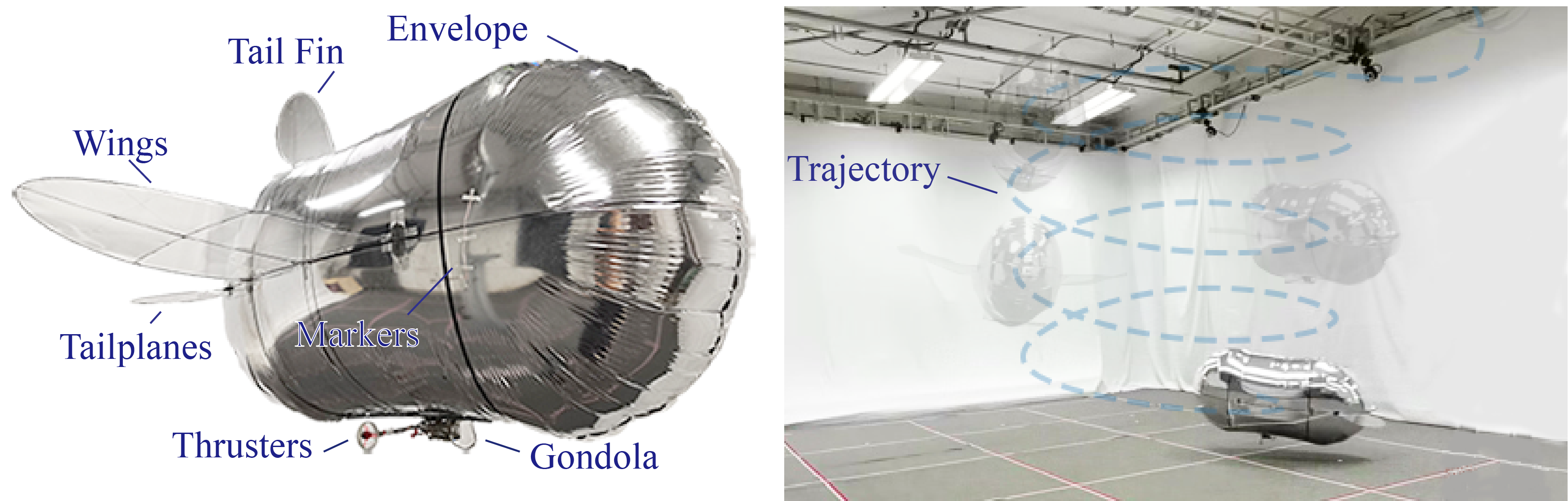}}
    \caption{Illustration of the RGBlimp prototype and the testing environment. The dashed line represents the trajectory of spiraling motion within the motion capture arena.}
    \label{fig:trajectory}
\end{figure}

Experiments were conducted using the RGBlimp prototype \cite{cheng2023rgblimp}, which consisted of a movable gondola equipped with a battery, a control unit, and a pair of propellers. In addition, the robot featured an envelope equipped with a pair of main wings, a tail fin, and twin tailplanes. The propellers provided differential propulsion for both forward motion and yaw adjustment, while the main wings were designed to enhance rolling stability and provide essential aerodynamic lift. Active markers affixed to the envelope were used for the motion capture system. An illustration of the prototype is presented in Fig. \ref{fig:trajectory}.

To validate the proposed modeling method, we conducted extensive experiments in an indoor environment. These tests took place in a 5.0m$\times$4.0m$\times$2.5m motion capture arena equipped with ten OptiTrack cameras. Operating at 60 Hz, the system provided motion data with an accuracy of 0.76mm RMS in positional estimation error. Each experiment involved adjusting differential propulsion by varying propeller thrust $F_l$ and $F_r$ as well as changing the flight posture by shifting the centroid of the gondola $\overline{\bm{r}}$. This led to spiral motions of various radii and speeds. Specifically, we varied the displacement $\Delta_{\overline{\bm{r}}_{x}}\text{[cm]}$ of the movable gondola at 6 different positions ranging from -1 to 4~cm, and adjusted the propeller thrust $(F_l,F_r)\text{[gf]}$ at 6 different configurations. The blimp failed to exhibit a stable spiral trajectory when $\Delta_{\overline{\bm{r}}_x} = -1/4$cm with certain force configurations. Ultimately, we selected a total of 30 effective spiral configurations, each comprising 4 independent trials, culminating in a dataset of 120 flight trajectories. Additionally, we conducted experiments using blimp's linear motion where the left and right thrusts were designed equal with configurations $(\Delta_{\overline{\bm{r}}_{x}},F_l,F_r)=(-1/0/1/2/3\text{cm},2.05\text{gf},2.05\text{gf})$, culminating in 20 trajectories. 
The average and maximum step lengths of the spiral motion are 2,030 and 7,450, respectively, while those of the linear motion are 468 and 580. 
\begin{figure*}[!t]
    \vspace{-3mm}
    \centerline{\includegraphics[]{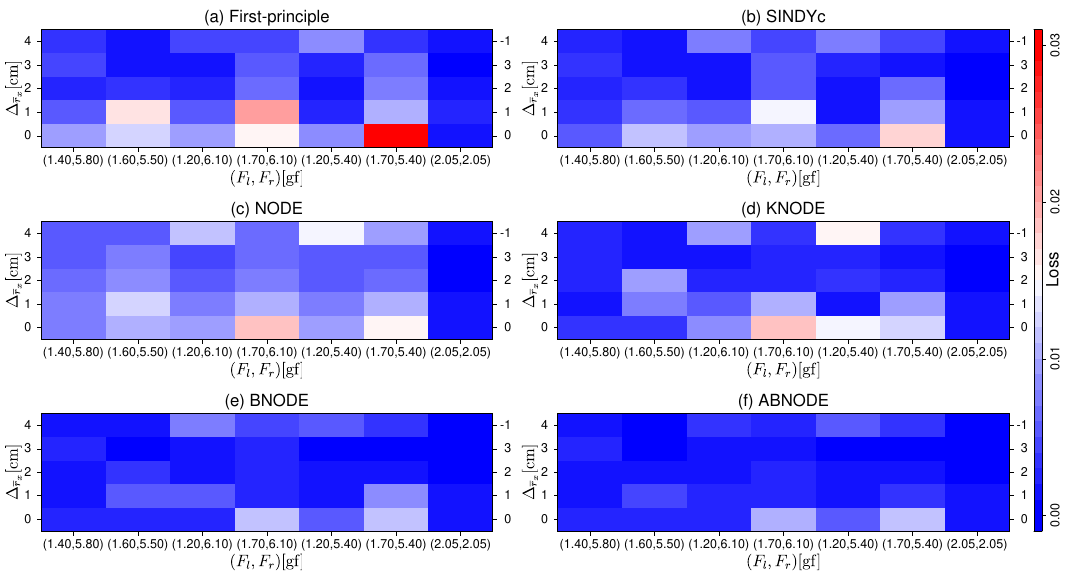}}
    \caption{Heat maps of the prediction loss for the first-principle, SINDYc, NODE, KNODE, BNODE and ABNODE models, tested with spiral and linear trajectories featuring various motion configurations. The vertical axis represents different displacements $\Delta_{\overline{\bm{r}}_{x}}$ of the gondola on the slide. The left vertical axis corresponds to the left two columns, and the right vertical axis corresponds to the right five columns. The horizontal axis represents different configurations of the propulsion thrust $F_l$ and $F_r$. }
    \label{fig:heat}
    \vspace{-3mm}
\end{figure*}
\subsection{Comparison Models}
We conducted a comparative analysis of the ABNODE method against five benchmark models including the first-principle, SINDYc, NODE, KNODE and BNODE. The first-principle model followed Eqs. \eqref{eq:full1} and \eqref{eq:full2}, with its imprecise parameters, such as aerodynamic coefficients, identified through solving the algebraic equations derived from quasi-steady-state flight experiments \cite{cheng2023rgblimp} while precise parameters, such as mass and dimensions, are physically measured with high confidence levels. 
SINDYc employed nonlinear candidate functions, including polynomials, sine and cosine functions, and their respective products (implemented via PySINDY \cite{desilva2020}) to model the residual dynamics of the robotic blimp.
NODE uses the standard MLPs for modeling, relying solely on data-driven approaches without incorporating domain knowledge about system dynamics. KNODE \cite{jiahao_knowledge-based_2021} incorporates the first-principle model (encompassing both dynamics and kinematics) into the NODE architecture, while BNODE and ABNODE incorporate only the dynamics. Unlike ABNODE, both BNODE and KNODE do not update the parameters of the first-principle model during training. A comparison of these models is detailed in Table \ref{Tab: model comparison}. 

\begin{table}[htbp]
\renewcommand{\arraystretch}{1.1}
\setlength{\tabcolsep}{3.2pt}
\begin{center}
\caption{Comparison between neural network-based models}
\label{Tab: model comparison}
\scalebox{1}{
\begin{tabular}{ccccc}
\toprule
  & NODE\cite{chen2018neural}         & KNODE\cite{jiahao_knowledge-based_2021}        & $\;$BNODE$\;$        & ABNODE   \\ \midrule
Kinematics       & \XSolidBrush & \Checkmark   & \XSolidBrush & \XSolidBrush \\
Dynamics         & \XSolidBrush & \Checkmark   & \Checkmark   & \Checkmark   \\
Auto-tuning     & \XSolidBrush & \XSolidBrush & \XSolidBrush & \Checkmark  
\\ \bottomrule
\end{tabular}
}
\end{center}
\end{table}

\subsection{Model Configuration}
Out of the 120 spiral and 20 linear experimental trajectories, we adopted a 3:1 train-test split. For every control input configuration, three trials were randomly selected for training the ABNODE and the four data-driven comparison models, while one spiraling trial was reserved for testing all six models. The network architecture of MLPs in the NODE model is consistent with that in the KNODE model for comparison purposes. The architecture includes a first layer with a dimension of $23\times 256$, a second layer of $256\times 64$, and a third layer of $64\times 12$. %The hyperbolic tangent (tanh) function is used as the activation function between layers. 
BNODE and ABNODE shared the same network structure as NODE and KNODE, except for their output layer, which had 6 dimensions corresponding to the dynamics output. 

Regarding parameter initialization, the uncertainty parameters in the first-principles model, including the aerodynamic force and moment coefficients and the damping coefficients, are initially obtained using traditional system identification methods via experiments, specifically polynomial regression methods not exceeding the fourth order. These identified uncertainty parameters, most likely not sufficiently accurate, are then used as the initial values for the first phase. In the second phase, the initial values for the network parameters are initialized using the Xavier uniform distribution (Glorot uniform initialization) method.

The 4th-order Runge Kutta and Adam are selected as the ODE solver and the optimizer, respectively. The step size was consistent with the experimental data, set at 1/60 seconds. The length of the prediction horizon $N$ corresponds to the length of the collected experimental data. Through trial and error, we set $N_1=N_2=10$ for the number of training epochs in each phase of ABNODE. For KNODE and BNODE, 10 epochs were also sufficient, while NODE required 100 epochs to converge. The performance of all models was evaluated using the loss function specified in Eq. \eqref{eq:allloss}. We used the weight matrix $\bm{W}$ for normalizing the data and eliminating dimensional units when calculating the loss. Data normalization is performed using Min-Max Scaling.

\section{Results And Analysis}
\label{section: Results}
\subsection{Accuracy}
We compared the prediction loss between ABNODE and other models. Figure \ref{fig:heat} illustrates the heat maps of the averaged prediction loss across all the test trials concerning the control input configurations. We observe that ABNODE consistently outperforms the first-principle model, achieving a significant $61.61\%$ reduction in loss across all control input configurations. Furthermore, ABNODE demonstrates improvements of $50.10\%$ and $19.53\%$ in loss reduction compared to SINDYc and BNODE, respectively.
NODE performs worse than the first-principle model, with an average loss that is 44.97\% higher. Compared to KNODE, BNODE exhibits a 39.21\% decrease in the average loss. Therefore, modeling only the dynamics is more effective than modeling both dynamics and kinematics. In subsequent experiments, we primarily compare the performance of ABNODE against the first-principle model, SINDYc, and BNODE.\par

Inferior model performance is observed with the configuration $(F_l, F_r)=(1.7\text{gf},5.4\text{gf})$ due to its longer experimental trajectories, averaging $41.2$ seconds for training and $41.8$ seconds for testing, compared to 31.8 and 34.0 seconds for other configurations. This results in higher cumulative errors from extended prediction horizons and more ODE iterations.\par

\begin{figure}[!h]
    \centerline{\includegraphics[]{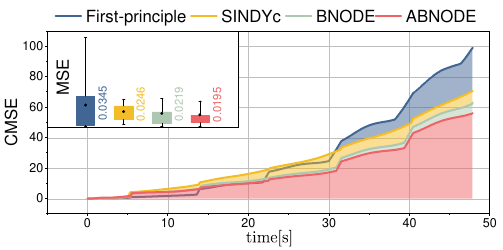}}
    \caption{CMSE of a single test trajectory, demonstrating the changing total loss over time. (Inset) MSE distributions across the entire trajectory, with black dots indicating the average MSE of the entire test trajectory.}
    \label{fig:result11}
\end{figure}

We adjusted the loss and designed two performance metrics, i.e., the time-dependent mean squared error (MSE) and time-dependent cumulative MSE (CMSE) of the prediction loss to delve deeper into the modeling accuracy over specific time periods. Let $\text{MSE}(t_i)$ denote the model prediction error at time $t_i$ and $\text{CMSE}(t_k)$ denote the sum of MSEs over $k$ steps, i.e., 
\begin{align}
  \text{MSE}(t_i)=&\frac{1}{n}\Vert \bm{W}(\bm{x}_\text{p}(t_i)-\bm{x}(t_i))\Vert_2^2  \label{eq:mse}\\
  \text{CMSE}(t_k)=&\sum_{i=2}^k\text{MSE}(t_i) \label{eq:cmse}
\end{align}
where $n$ represents the dimension of system states, and $\bm{x}(t_i)$ and $\bm{x}_\text{p}(t_i)$ represent the experimental data and model prediction of the states, respectively. The weight matrix $\bm{W}$ in Eq. (\ref{eq:mse}) is the same as in Eq. (\ref{eq:allloss}). As an illustration, Fig.~\ref{fig:result11} showcases the trajectory of CMSE and the distributions of MSE for the configuration $(\Delta_{\overline{\bm{r}}_{x}},F_l,F_r)=(0\text{cm},1.2\text{gf},5.4\text{gf})$. ABNODE exhibits the highest modeling accuracy compared to other models, evident from its smallest average MSE, indicating its temporal consistency and robustness. 

\subsection{Generalizability}
We evaluated the generalization capability of the four modeling methods using the spiraling experimental dataset. To do so, we trained models using one control input configuration and tested them on neighboring configurations For instance, the configuration $(\Delta_{\overline{\bm{r}}_{x}},F_l,F_r)=(1\text{cm},1.2\text{gf},6.1\text{gf})$ had four neighboring configurations, as depicted in Fig. \ref{fig:heat}, including $(\Delta_{\overline{\bm{r}}_{x}},F_l,F_r)=(1\text{cm},1.6\text{gf},5.5\text{gf})$, $(1\text{cm},1.7\text{gf},6.1\text{gf})$, $(0\text{cm},1.2\text{gf},6.1\text{gf})$, and $(2\text{cm},1.2\text{gf},6.1\text{gf})$. We excluded the edge configurations and selected 12 setups, each with four neighboring configurations, resulting in a total of 48 configurations for evaluating the model generalization capability.

\begin{figure}[!htbp]
    \centerline{\includegraphics[]{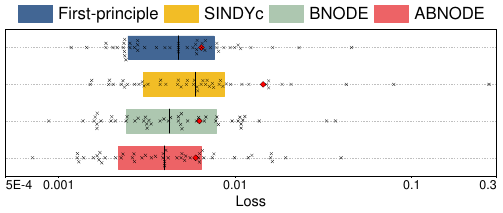}}
    \caption{Boxplot of the model prediction loss (Eq. \ref{eq:allloss}) across four comparison models. The box represents the interquartile range (IQR), spanning from the first quartile (Q1) to the third quartile (Q3). The line inside the box represents the median while the red diamond dot represents the mean. (Overlap) Beeswarm scatter plot of loss values. Black crosses represent the 48 sets of test results corresponding to each model. The horizontal axis employs a logarithmic scale.}
    \label{fig:boxLine}
\end{figure}

Figure \ref{fig:boxLine} illustrates the aggregated statistical results of the model prediction loss over all generalizability tests. ABNODE outperforms the first-principle model, SINDYc, and KNODE in terms of the averaged prediction loss by $7.17\%$, $58.51\%$, $4.80\%$, respectively, indicating superior generalization capability compared to other benchmark models. Regarding the interquartile range (IQR), ABNODE demonstrates the best performance, being lower than those of the first-principle model, SINDYc, and KNODE by $17.32\%$, $26.19\%$, and $21.93\%$, respectively. Meanwhile, ABNODE possesses the lowest standard deviation, indicating the highest consistency across experiments. 

\subsection{Performance Analysis on Parameter AutoTuning}
We evaluated the performance improvement achieved through parameter auto-tuning in the first phase of model training. This analysis involved quantifying the relative loss reductions of the individual phases using three performance indices, i.e.,
\begin{align}
    \lambda_1=&\frac{\mathcal{L}_{\rm{phy}}(\bm{\eta})-\mathcal{L}_{\rm{phase1}}(\bm{\eta}^{\star})}{\mathcal{L}_{\rm{phy}}(\bm{\eta})} \label{eq:lambda1} \\
    \lambda_2=&\frac{\mathcal{L}_{\rm{BNODE}}(\hat{\bm{\theta}}^{\star})-\mathcal{L}_{\rm{ABNODE}}(\bm{\eta}^{\star},\bm{\theta}^{\star})}{\mathcal{L}_{\rm{BNODE}}(\hat{\bm{\theta}}^{\star})} \label{eq:lambda2} \\
    \lambda_3=&\frac{\mathcal{L}_{\rm{phy}}(\bm{\eta})-\mathcal{L}_{\rm{ABNODE}}(\bm{\eta}^{\star},\bm{\theta}^{\star})}{\mathcal{L}_{\rm{phy}}(\bm{\eta}))} \label{eq:lambda3} 
\end{align}
where $\lambda_1$ is the loss reduction of the first phase (with physical parameter auto-tuning) with respect to the first-principle model (without physical parameter auto-tuning). Similarly, $\lambda_2$ and $\lambda_3$ represent the loss reductions of the second phase (after physical parameter auto-tuning in the first phase) with respect to the BNODE and first-principle models (neither with physical parameter auto-tuning), respectively. \par

\begin{figure}[!htbp]
    \centerline{\includegraphics[width=9cm]{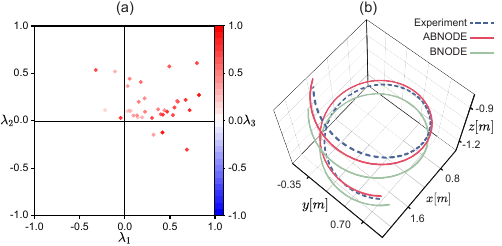}}
    \caption{Analysis of parameter auto-tuning. (a) The scatter plot of three performance indices $\lambda_1$, $\lambda_2$, and $\lambda_3$, demonstrating that the integration of parameter auto-tuning in the first phase contributes to enhanced modeling accuracy in both the first-principle and neural network modules. (b) An example spiral trajectory for $(\Delta_{\overline{\bm{r}}_{x}},F_l,F_r)=(2\text{cm},1.4\text{gf},5.8\text{gf})$. The predicted trajectory of ABNODE closely aligns with the experimental result, showcasing its superior performance compared to BNODE, which lacks physical parameter auto-tuning. }
    \label{fig:scatter_fig}
\end{figure}

Figure \ref{fig:scatter_fig} (a) shows the scatter plot of $\lambda_1$, $\lambda_2$, and $\lambda_3$. Each data point corresponds to one individual experimental trial. The clusterxing of points primarily lies in the first quadrant, indicating a positive correlation between the performance gain of parameter auto-tuning in the first phase and that in the second phase. Additionally, the intensity of the red color reflects the degree of performance gain achieved by the entire two-phase model training over the first-principle model, with darker shades indicating higher loss reduction. In the scatter plot, darker red dots are mainly distributed in the first quadrant. This observation suggests that the integration of parameter auto-tuning in the first phase contributes to enhanced modeling accuracy in both the first-principle and neural network modules.\par

To visually demonstrate the modeling improvement achieved through parameter auto-tuning, Fig. \ref{fig:scatter_fig} (b) illustrates an example spiraling flight trajectory in three-dimensional space. From this visualization, we observe that the model prediction of ABNODE aligns closely with the experimental trajectory, demonstrating a significant improvement compared to BNODE which lacks parameter auto-tuning.

\subsection{\texorpdfstring{Discussions on ABNODE Models}{}}
\subsubsection{\texorpdfstring{Adaptability to prediction horizons and time steps}{}}
All models were trained with a prediction time step of 1/60\,s in aforementioned experiments. To further investigate, we conducted experiments and evaluated the ABNODE models on a test dataset with sequence lengths exceeding 45 seconds. These experiments employed varying prediction horizons and time steps. Specifically, we evaluated 10 ABNODE models with time steps set at 2/60\,s, 3/60\,s, 4/60\,s, 5/60\,s, and prediction horizons set at 30\,s, 35\,s, 40\,s, 45\,s. Consequently, varing prediction time steps resulted in different data lengths under consistent total time lengths. Additionally, we conducted experiments with mixed prediction time steps of 1/60\,s, 2/60\,s, and 3/60\,s in a cyclical pattern. Table \ref{loss_percentage} shows the average percentage change in loss compared to a time step of 1/60\,s across varying time steps under the same corresponding prediction horizons. We observe that while different prediction time steps do result in variations in prediction loss, the average loss increase remains below 0.4\% across all selected prediction horizons, indicating that the ABNODE model has satisfactory adaptability to varying time steps.

\begin{table}[htbp]
\renewcommand{\arraystretch}{1.05}
\setlength{\tabcolsep}{9.5pt}
\begin{center}
\caption{Average loss percent change of ABNODE models under various time steps compared to the 1/60\,s time step at selected prediction horizons. All the values have a unit of percent (\%).}
\label{loss_percentage}
\scalebox{0.95}{
\begin{tabular}{cc|ccccc}
\toprule
& & \multicolumn{4}{c}{\textbf{Prediction Horizon}}\\
   &   & 30\,s & 35\,s & 40\,s & 45\,s \\ \midrule
\multirow{6}{*}{\rotatebox{90}{\textbf{Time Step}}} & mixed & 0.02  & -0.15     & -0.22    & -0.18    \\ 
& 2/60\,s & $\bm{0.13}$  & $\bm{0.17}$   & 0.06      & 0.06     \\
& 3/60\,s & -0.49   & -0.43       & -0.42   & -0.13  \\
& 4/60\,s & -0.14   & 0.05       & $\bm{0.12}$    & -0.13     \\
& 5/60\,s & -0.33    & -0.38    & 0.01    & $\bm{0.34}$      \\ \bottomrule
\end{tabular}
}
\end{center}
\end{table}
\subsubsection{\texorpdfstring{Robustness to parameter initial values}{}}
We conducted experiments to investigate the model's robustness to the initial values of the physical parameters in the first phase of training. The experiments employed data from six different spiral motion configurations with $\Delta_{\overline{\bm{r}}_{x}}=2\text{cm}$. Each test involves perturbing $\bm{\eta}_{K_{i}}$ where $i=1,2,3$ with four multiplicative levels: $0.0$, $0.5$, $1.5$, and $2.0$, representing -100\%, -50\%, +50\%, and +100\% perturbation, respectively. This results in eighteen models trained at each perturbation level across the six configurations. The average losses on the test trajectories are detailed in Table \ref{Tab: initial_disturb}, which also includes the losses of models without perturbations, labeled as $1.0$ for comparison. Comparing these variations to the original parameters, we observed a substantial increase in the loss of the first-principle model. Specifically, at a disturbance of $200\%$, the loss growth rate of the first-principle model was approximately 21 times greater than that of the ABNODE model, demonstrating ABNODE's superior robust performance to inaccurate initial parameter values. 

\begin{table}[htbp]
\renewcommand{\arraystretch}{1.1}
\setlength{\tabcolsep}{8pt}
\begin{center}
\caption{Average loss comparison between various levels of multiplicative perturbation in the initial values of physical parameters using first principle and ABNODE models ($10^{-2}$).}
\label{Tab: initial_disturb}
\scalebox{0.95}{
\begin{tabular}{c|ccccc}
\toprule
 & \multicolumn{5}{c}{\textbf{multiplicative levels}}\\
  & 0.0     & 0.5      & 1.0      & 1.5      & 2.0    \\ \midrule
First-principle & 3.73  & 1.30 & 0.38 & 1.46 & 3.46 \\
ABNODE          & 0.32 & 0.30 & 0.15 & 0.20 & 0.21 \\ \bottomrule
\end{tabular}
}
\end{center}
\end{table}

\section{Conclusion}
\label{section: Conclusion}
This letter proposed the ABNODE model that consisted of a first-principle module with physical parameter auto-tuning and a neural network module for modeling residual dynamics in robotic blimps. Extensive spiraling experiments were conducted, the results of which demonstrated improved modeling accuracy and generalizability compared to three baseline models including the first-principle, SINDYc, and BNODE. \par
In future work, we plan to extend the application scope of the ABNODE model to accommodate a wider range of motions executed by robotic blimps, e.g., large attack angles, stall maneuvers, and high velocities. Additionally, we will investigate feedback control designs built upon the ABNODE model for agile and complex flight tasks by adding environmental interference in real-world application. 

\section{Acknowledgement}
The authors would like to thank Prof. Zhongkui Li for his tremendous help in the motion capture experiment.

% \newpage
\bibliographystyle{IEEEtran}

% \clearpage
% \renewcommand{\headrulewidth}{0pt}
% \fancyhead[C]{}
\bibliography{references.bib}

% \bibliographystyle{IEEEtran}
% \bibliography{references.bib}

\end{document}